\def\paperID{44523}
\def\confName{CVPR}
\def\confYear{2026}

\def\paperTitle{Any2Any 3D Diffusion Models with Knowledge Transfer: \\ \textit{A Radiotherapy Planning Study}}

\def\authorBlock{
Yuhan Wang$^{1,2,\dagger}$ \quad
Zihan Li$^{2,3,\dagger}$ \quad
Han Liu$^{2}$ \quad
Simon Arberet$^{2}$ \quad
Martin Kraus$^{2}$ \\
Yuyin Zhou$^{1}$ \quad
Florin-Cristian Ghesu$^{2}$ \quad
Dorin Comaniciu$^{2}$ \quad
Ali Kamen$^{2}$ \quad
Riqiang Gao$^{2,}$\thanks{Corresponding author: riqiang.gao@siemens-healthineers.com} \\
$^{1}$UC Santa Cruz \quad
$^{2}$Siemens Healthineers \quad
$^{3}$University of Washington \quad
$^{\dagger}$Equal contribution
}

\newif\ifreview 
\newif\ifarxiv \newcommand{\arxiv}{\arxivtrue}
\newif\ifcamera 
\newif\ifrebuttal 

\arxiv  

\pdfoutput=1
\documentclass[10pt,twocolumn,letterpaper]{article}
\ifreview \usepackage[review]{cvpr} \fi
\ifarxiv \usepackage[pagenumbers]{cvpr} \fi
\ifrebuttal \usepackage[rebuttal]{cvpr} \fi
\ifcamera \usepackage{cvpr} \fi

\usepackage{graphicx}	
\usepackage{amsmath}	
\usepackage{amssymb}	
\usepackage{booktabs}
\usepackage{times}
\usepackage{microtype}
\usepackage{epsfig}
\usepackage{caption}
\usepackage{float}
\usepackage{placeins}
\usepackage{color, colortbl}
\usepackage{stfloats}
\usepackage{enumitem}
\usepackage{tabularx}
\usepackage{xstring}
\usepackage{multirow}
\usepackage{xspace}
\usepackage{url}
\usepackage{subcaption}
\usepackage[most]{tcolorbox}
\usepackage[hang,flushmargin]{footmisc}
\usepackage[dvipsnames]{xcolor}

\usepackage{tocloft}
\ifcamera \usepackage[accsupp]{axessibility} \fi

% ...

\definecolor{demphcolor}{RGB}{90,90,90}

\newcommand{\nbf}[1]{{\noindent \textbf{#1.}}}

\ifarxiv  \fi

\newcommand{\R}[1]{{%
    \textbf{%
        \ifstrequal{#1}{1}{\textcolor{red}{R#1}}{%
        \ifstrequal{#1}{2}{\textcolor{blue}{R#1}}{%
        \ifstrequal{#1}{3}{\textcolor{magenta}{R#1}}{%
        \ifstrequal{#1}{4}{\textcolor{teal}{R#1}}{%
                           \textcolor{cyan}{R#1}%
        }}}}%
    }%
}}

\usepackage{xr-hyper}

\makeatletter
\newcommand*{\addFileDependency}[1]{
  \typeout{(#1)}
  \@addtofilelist{#1}
  \IfFileExists{#1}{}{\typeout{No file #1.}}
}

\makeatother
\newcommand*{\myexternaldocument}[1]{
    \externaldocument{#1}
    \addFileDependency{#1.tex}
    \addFileDependency{#1.aux}
}

\definecolor{cvprblue}{rgb}{0.21,0.49,0.74}
\usepackage[pagebackref,breaklinks,colorlinks,allcolors=cvprblue]{hyperref}
\usepackage[capitalize]{cleveref}
\crefname{section}{Sec.}{Secs.}
\crefname{table}{Table}{Tables}
\crefname{figure}{Fig.}{Figs.}

\ifarxiv \crefname{appendix}{App.}{Apps.}
\else \crefname{appendix}{Suppl.}{Suppls.} \fi

\unless\ifarxiv \myexternaldocument{_supplementary} \fi

\begin{document}
\ifarxiv \addtocontents{toc}{\protect\setcounter{tocdepth}{-1}} \fi
\title{\paperTitle}
\author{\authorBlock}
\maketitle

\begin{abstract}
Voxel-wise dose prediction is a critical yet challenging task in practical radiotherapy (RT) planning, as bespoke models trained from scratch often struggle to generalize across diverse clinical settings. Meanwhile, generative models trained on billion-scale datasets from vision domains have achieved impressive performance. Herein, we propose \textbf{DiffKT3D}, a unified Any2Any 3D diffusion framework that leverages prior knowledge from pretrained video diffusion models for efficient and clinically meaningful dose prediction. To enable flexible conditioning across multiple clinical modalities (CT, anatomical structures, body, beam settings, etc.), we introduce an Any2Any conditional paradigm utilizing modality-specific embeddings without cross-attention overhead. Further, we design a novel reinforcement learning (RL) post-training mechanism guided by a clinically-informed Scorecard explicitly tailored to institutional treatment preferences. Compared with winner of GDP–HMM challenge, DiffKT3D sets a new state-of-the-art in dose prediction by reducing voxel-level MAE from 2.07 to 1.93. In addition, DiffKT3D achieves superior image quality and preference match. These results demonstrate that transferring diffusion priors via modality-aware conditioning and clinically aligned RL post-training can provide a robust and generalizable solution for RT planning across various clinical scenarios.

\end{abstract}
\section{Introduction}
\label{sec:intro}

Radiotherapy (RT) is one of the most commonly used cancer treatments. Dose prediction (DP) is a prominent AI application in RT planning, aiming to generate a 3D dose distribution from patient data and machine configurations by learning from large-scale historical, deliverable treatment solutions \cite{Babier2020OpenKBP,Gao2023FlexibleCmGAN,gao2025automating,kui2024review}. DP task typically includes a planning CT scan and delineated structures e.g., planning target volumes (PTVs), organs at risk (OARs) as input, and reference 3D RT dose as targets. 
Dose predictors have wide applications in RT pipeline, including planning optimization \cite{Babier2022OpenKBPOpt,AAPM2025GDPHMM}, fluence prediction \cite{Wang2020FluenceTherapy}, leaf sequencing \cite{gao2024multi}, and quality improvement \cite{Gronberg2023DeepPlans}. 
Towards precise and personalized prediction, additional conditions such as beam geometries and prescriptions can also be incorporated. 

DP task has been historically addressed as a supervised regression problem, training models to minimize voxel-wise errors (e.g., MAE, MSE) between predicted and reference dose \cite{Babier2020OpenKBP,gao2025automating}. Recent advances in generative modeling have opened new avenues for DP. Generative models such as GANs \cite{kearney2020dosegan,Gao2023FlexibleCmGAN}, and diffusion models \cite{Feng2023DiffDP,Zhang2023DoseDiff} have been explored for dose prediction, showing promise in capturing complex dose distributions and improving prediction quality. Existing work primarily are trained from scratch and without post-training for e.g., clinical alignment.
\begin{figure}[t]
    \centering
    \includegraphics[width=0.99\linewidth]{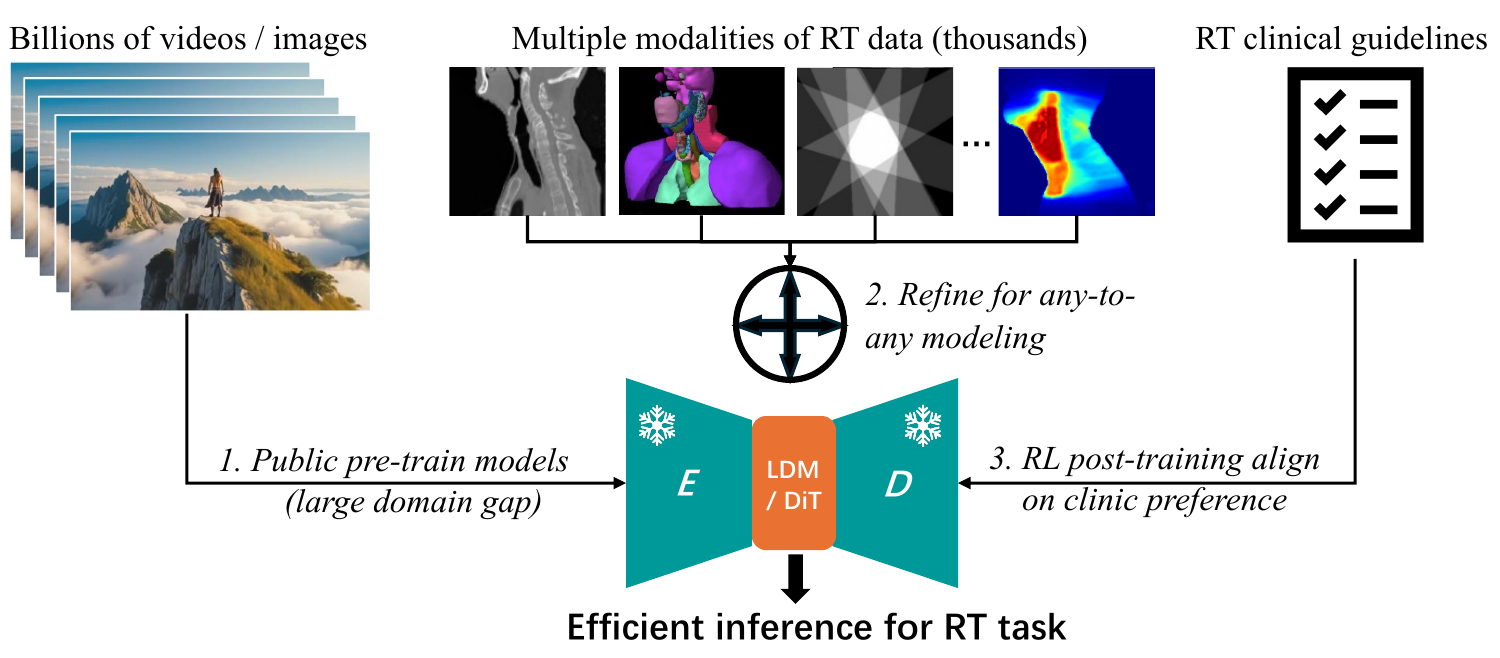}
    \vspace{-0.1in}
    \caption{Illustration of the proposed DiffKT3D. We first transfer priors from diffusion models pretrained on large-scale public video or CT data, despite a substantial domain gap to radiotherapy dose prediction. These backbones are then adapted to heterogeneous RT modalities with relatively limited data, followed by RL post-training driven by guideline-derived clinical Scorecards to better align predictions with institutional planning preferences.}
    \label{fig:intro}
    \vspace{-0.2in}
\end{figure}

Concurrently, foundational models have been promising in natural language processing, computer vision, and medical imaging \cite{Oquab2023DINOv2,Radford2021CLIP,Azad2023FMMISurvey}. Those primarily trained with self-supervised learning or unrelated tasks but show to transfer effectively to target tasks via light adaptation or even training-free \cite{Song2024DINOReg,Wu2023MedSA,Huang2024MedAD}. More surprisingly, recent studies show that \emph{general-domain} foundational models pretrained on natural images can also transfer effectively to medical imaging tasks \cite{Song2024DINOReg,Huang2024MedAD,Guo2025Text2CT,Zhang2023BiomedCLIP,Lin2023PMCClip}, despite the significant domain gap. However, previous work mainly on the feature extraction backbone rather than generation, this motivates our question about first type knowledge transfer: \emph{Can 3D diffusion prior knowledge trained on a distant source domain help improve target-domain generation?}

 User preference is another critical consideration in generative AI \cite{Ouyang2022InstructGPT,Rafailov2023DPO}. It is particularly important in RT because multi-disciplinary team (oncologists, physicists, dosimetrists) collaboratively design treatment plans tailored to individual patients, and different institutions follow slightly or largely different protocols \cite{Bentzen2010QUANTEC,Ezzell2009TG119}. Post-training with reinforcement learning (RL) has emerged as a powerful paradigm to align generative models with user preferences in NLP and CV \cite{Black2023DDPO,Wallace2024DiffusionDPO,Yang2024D3PO,Xu2023ImageReward,Kirstain2023PickaPic}.  However, its application in medical imaging, especially RT dose planning, remains underexplored. User preference in RT dose planning is multifaceted, involving complex trade‑offs between PTV coverage and OAR sparing usually reflecting in institutional protocols. This motivates our question about second type knowledge transfer: \emph{Can we align diffusion generations with clinical preferences via RL post‑training?}

As illustrated in Figure \ref{fig:intro}, we propose \textit{\textbf{DiffKT3D}}, a novel \textbf{3D} \textbf{Diff}usion model framework addressing two critical \textbf{K}nowledge \textbf{T}ransfer questions. Three main contributions of DiffKT3D are outlined below. 

\textit{\textbf{First}}, we adapt advanced 3D diffusion models pretrained on natural videos (Wan 2.1 \cite{wan2025wan}) or CT data (MAISI \cite{zhao2025maisi}) and fine-tune them slightly for dose prediction. Despite substantial domain gaps, this approach yields notable improvements in both accuracy and efficiency. Moreover, the benefit of diffusion-based transfer becomes even more pronounced when the domain gap is smaller, demonstrating significantly stronger generalization than regression-based models. 

\textit{\textbf{Second}}, considering \textbf{(1)} the heterogeneous nature of DP data which may involve seven modalities \{CT, PTV, OAR, body, beam plate, angle plate, dose\}, and \textbf{(2)} our findings on cross domain / modal diffusion knowledge transfer, we introduce a modality- and role-aware \textbf{Any2Any} conditioning scheme. In this framework, \textit{any} modalities can serve as the target while the remaining modalities act as conditions. This design enables flexible handling of variable input combinations and supports  dose prediction in diverse scenarios. 

\textit{\textbf{Third}}, we introduce a RL post-train, termed as \textbf{\textit{ScardNFT}}, with a new rewarding mechanism based on planning preferences captured by \textbf{S}core\textbf{card}. This method is inspired by success of Diffusion\textbf{NFT} \cite{Zheng2025DiffusionNFT} in text-to-image alignment and tailored for clinical-guided refinement. 

We experiment DiffKT3D on over 8,000 plans from the GDP–HMM Grand Challenge \cite{gao2025automating} (head-and-neck and lung) and prostate plans from the REQUITE patients \cite{seibold2019requite}. Our method achieves substantial MAE gains over top challenge solutions and shows improved image quality and preference alignment. Beyond diffusion priors from non-RT domains, DiffKT3D demonstrates strong transferability: models pretrained on head-and-neck and lung data adapt quickly to prostate cancer with minimal fine-tuning, offering clear practical benefits for efficiently supporting new disease sites with limited computational resources. Our study conducts extensive validation in RT planning context, core ideas of DiffKT3D are broadly applicable to other generative tasks.

\section{Related Work}

\nbf{RT Dose Prediction}
Voxel-wise dose prediction has gained popularity in recent years due to advances in deep learning. Inspired by the success in image segmentation, Convolutional UNet \cite{ronneberger2015u} and its variants including ResUNet \cite{diakogiannis2020resunet}, H-DenseUNet \cite{li2018h}, and MedNeXt \cite{roy2023mednext} have been used for different cancer sites including head-and-neck \cite{nguyen20193d,Liu2021TechnicalRadiotherapy,Soomro2021DeepDoseNet,Wang2022DeepDecomposition}, lung \cite{Gao2023FlexibleCmGAN,Barragan-Montero2019Three-dimensionalConfigurations,jhanwar2022domain}, prostate \cite{kearney2020dosegan,nguyen2020incorporating,Kearney2018DoseNet:Networks}, esophageal \cite{Zhang2020PredictingConvolutions,Babier2020Knowledge-basedNetworks}. 
Although many studies use regression losses such as L1 or L2, researchers \cite{kearney2020dosegan,Gao2023FlexibleCmGAN,Feng2023DiffDP,Zhang2023DoseDiff} have also explored generative methods, including GANs \cite{goodfellow2014generative} and diffusion models \cite{ho2020denoising}, to improve image quality. Across challenges like OpenKBP \cite{Babier2020OpenKBP} and GDP-HMM \cite{gao2025automating}, there is growing emphasis on generalizable models that handle multiple contexts rather than highly specialized ones. In computer vision, diffusion models often outperform GANs in complex scenarios, but state-of-the-art diffusion methods are data-hungry, computationally expensive, and mostly designed for 2D. Existing work \cite{Feng2023DiffDP,Zhang2023DoseDiff} trains slice-wise diffusion models, which struggle with spatial consistency across slices, highlighting needs of efficient 3D diffusion models for dose prediction.

\nbf{Diffusion Priors and Any2Any Generation}
Diffusion models have achieved notable success in conditional generation tasks across various modalities, initially excelling in image synthesis and later adapted for dense prediction tasks like depth estimation and segmentation \cite{Ke2023Marigold}. Recent frameworks have further extended diffusion models to Any2Any generation, enabling unified conditional generation for arbitrary modality pairs, exemplified by Versatile Diffusion \cite{Xu2022VersatileDiffusion}, UniDiffuser \cite{Bao2023UniDiffuser}, CoDi \cite{Tang2023CoDi}, OmniGen \cite{Xiao2024OmniGen}, and OmniFlow \cite{Li2024OmniFlow}. Techniques like ControlNet \cite{Zhang2023ControlNet}, T2I-Adapter \cite{Mou2023T2IAdapter}, and MultiDiffusion \cite{BarTal2023MultiDiffusion} further facilitate flexible integration of diverse conditions, including edge maps and segmentation masks. Additionally, instruction-tuned models such as PixWizard \cite{Lin2024PixWizard} and joint generation-understanding models like JoDI \cite{Xu2025JoDI} enhance practical applicability. Inspired by these methodological advances, we adapt diffusion priors to radiotherapy dose prediction, leveraging their flexibility and cross-modal generalization for robust clinical outcomes.

\nbf{Post-training for Diffusion Models}
Standard diffusion models typically optimize voxel-level losses (e.g., MSE or MAE), which often misalign with nuanced clinical objectives. To address complex, preference-driven tasks, recent methods such as DDPO \cite{Black2023DDPO}, Diffusion-DPO \cite{Wallace2024DiffusionDPO,Yang2024D3PO}, and DiffusionNFT \cite{Zheng2025DiffusionNFT} apply reinforcement learning (RL) strategies for preference-based fine-tuning. However, their clinical applicability remains underexplored, as general-purpose preference frameworks like ImageReward \cite{Xu2023ImageReward}, HPS \cite{Wu2023HPS}, PickaPic \cite{Kirstain2023PickaPic}, and MPS \cite{Zhang2024MPS} lack necessary interpretability. Radiotherapy dose prediction involves complex trade-offs encoded in institutional guidelines, motivating our clinically-informed RL-based post-training strategy. We introduce a Scorecard reward mechanism explicitly encoding clinical metrics, guiding diffusion generations toward institutional preferences and enhancing clinical acceptability.

\begin{figure*}[t]
  \centering
  \includegraphics[width=\textwidth]{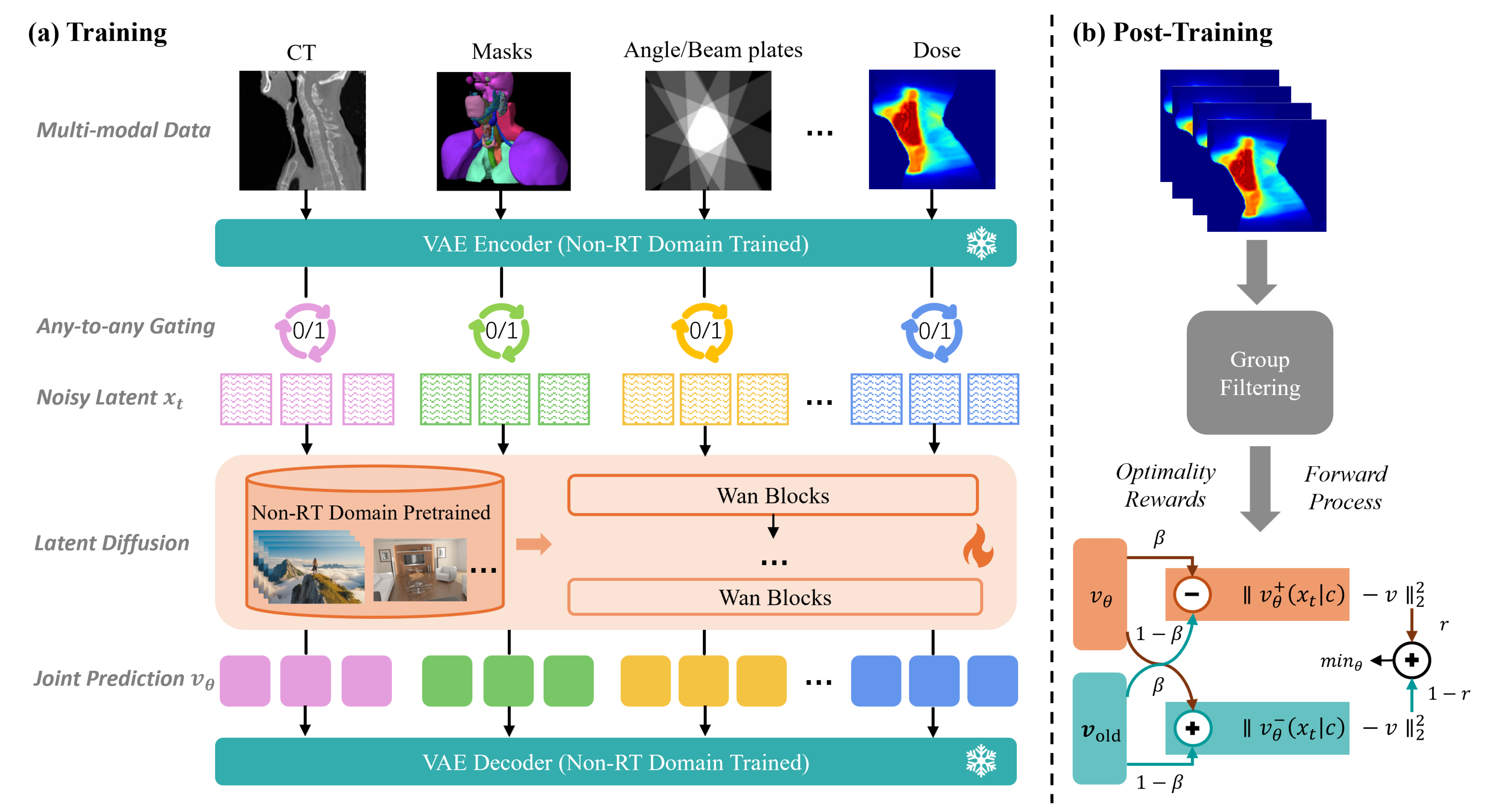}
  \vspace{-1.5em}
  \caption{Training mechanism for DiffKT3D. The multi-modal data first pass through the VAE encoder to obtain latent features. With the Any2Any gating mechanism, each modality is randomly assigned as either a condition or a target. Conditional modalities are independently encoded into patch tokens, while target modalities are combined with latent noise $x_t$. Each token is annotated with a domain embedding (indicating its modality) and a role embedding (distinguishing targets from conditions). The DiT jointly attends to the clean condition tokens and the noised target tokens, predicting the noise parameterization ($v_\theta$) for the selected target modality. The VAE and DiT are pretrained on videos, and only the DiT blocks are fine-tuned. During post-training, we convert a clinically informed Scorecard into an RL reward, improving clinical preference alignment while maintaining voxel-level fidelity.}
  \label{fig:arch}
  \vspace{-1.5em}
\end{figure*}
\section{Method}
\label{sec:method}
\subsection{Problem Description and Motivation}
\label{sec:method:setup}
RT planning involves designing a treatment strategy that precisely delivers radiation to the target region while minimizing exposure to nearby healthy tissues and organs.
Dose prediction aims to generate accurate 3D dose distributions from diverse multimodal inputs (as shown in Figure~\ref{fig:intro}), often under inconsistent or incomplete contouring of the PTVs and OARs.
Because planning protocols vary across institutions, preferences in plan quality also differ.
These variations can be captured using clinical Scorecards \cite{magliari2025hn,varian2024han,varian2024lungconv}, which quantitatively evaluate and compare treatment plans.
Consequently, \textit{generalizable dose prediction can be framed as a 3D generation problem conditioned on varied inputs and guided by preference-oriented objectives.}

\vspace{0.05in}
\nbf{Motivation for Methodology}
Diffusion-based generative models have achieved remarkable success in computer vision, typically trained on billions of samples.
In contrast, most dose prediction models rely on only hundreds or thousands of cases due to the limited availability of RT data.
The DINO family has demonstrated strong feature extraction capabilities that transfer well to medical imaging even when trained on natural images, yet our work focuses on generation.
We therefore explore leveraging large-scale diffusion priors from non-RT domains for dose prediction, forming our first methodological contribution.

After establishing that diffusion-based generative knowledge can effectively transfer across domains and modalities (\textit{first contribution}), we further aim to enhance performance within the RT domain, which involves diverse multimodal inputs.
To this end, we formulate dose prediction as an \textit{Any2Any} conditional generation task, enabling flexible selection of any target modality $\tau$ from the set $\mathcal{M}{=}\{\texttt{ct},\texttt{ptv},\texttt{oar},\texttt{body},\texttt{beam},\texttt{angle},\texttt{dose}\}$, conditioned on the remaining available inputs $C{=}\{x_0^{(m)} \mid m\in S\subseteq\mathcal{M}\setminus\{\tau\}\}$.
The target modality at the $t$-th step of the forward diffusion process becomes:
\vspace{-0.05in}
\begin{equation}
x_t^{(\tau)} = \alpha_t x_0^{(\tau)} + \sigma_t \varepsilon,\quad 
\varepsilon\sim\mathcal{N}(0,\mathbf{I}),~ t\sim\mathcal{U}(0,1),
\label{eq:forward}
\end{equation}
where $x_0$ denotes the clean data, and $\mathcal{N}$ and $\mathcal{U}$ represent Gaussian and uniform distributions, respectively.
The coefficients $\alpha_t$ and $\sigma_t$ follow a standard variance-preserving noise schedule (with $\alpha_t^2+\sigma_t^2{=}1$), and discrete diffusion steps are linearly rescaled to $t\in[0,1]$ for notational simplicity.
This formulation allows any modality in $\mathcal{M}$ to serve as the generation target, facilitating cross-modality diffusion knowledge transfer while mitigating overfitting in low-data regimes.

In addition, reinforcement learning (RL) has achieved notable success in LLM post-training and text-to-image alignment, both of which share a common goal with RT clinical Scorecards: capturing user or expert preferences.
Motivated by this similarity, we adapt the state-of-the-art DiffusionNFT framework by reformulating the clinically evaluation-based Scorecard into an RL reward function for post-training.
This provides an additional safeguard for out-of-distribution cases by explicitly aligning the generative model with clinically preferred trade-offs.

Figure~\ref{fig:arch} summarizes our approach, with detailed model structure provided in Supplementary \ref{app:struct}, and method details presented in the following subsections.

\subsection{Any2Any DiT framework}
\label{sec:method:arch}

\nbf{Patchify Head}
Our diffusion process operates entirely within a shared VAE latent space. For each modality $m$, raw input volumes are first encoded into latent representations by a pretrained, frozen VAE encoder. We then reuse the 3D patch embedding block from the Wan DiT backbone and extend it to be modality-specific: a lightweight modality-specific patch embedding $\mathrm{PE}_m$, implemented as a compact 3D convolution with the same structure as the original DiT patch embed, projects these latent grids (or their noised versions $x_t^{(\tau)}$ for target modalities) into tokens with hidden dimension $D$. After diffusion in token space, predictions are decoded back to the original voxel space via the VAE decoder. This design keeps the DiT backbone architecture unchanged while delegating modality handling to the patch-embedding blocks, allowing the backbone to operate in a unified latent token space.

\nbf{Role-aware conditioning and AdaLayerNorm}
We use a single binary role embedding $e^{\text{role}} \in \{e^{\text{tar}}, e^{\text{cond}}\}$ to tag each token as either the noised prediction target or a clean condition. To inject conditioning into the backbone, we construct a global conditioning code $e_C$ directly from the conditional role embedding $e^{\text{cond}}$ and simply add this vector to the original timestep embedding $e_t$, yielding a fused signal $\tilde{e}_t = e_t + e_C$. Following Wan, all transformer blocks share one AdaLayerNorm modulation network, and $\tilde{e}_t$ is the only input to this shared AdaLayerNorm in every block. This scheme keeps conditioning as a simple additive modulation on the timestep embedding, reuses Wan’s AdaLayerNorm parameters, introduces no additional pooling or cross-attention modules, and allows the backbone to jointly model targets and conditions using full self-attention.

\nbf{Slot-aware 4D RoPE positional embeddings}
To explicitly encode token origins (dose vs.\ conditions) while preserving precise 3D spatial relationships, we extend standard 3D RoPE with an additional slot axis, yielding a 4D RoPE.
Tokens are arranged in slot-major order, where each slot index $S$ corresponds to one modality (the target dose or a particular condition).

For each attention head, we split the channel dimension $d$ into four sub-dimensions assigned to the slot and the three spatial axes:
\begin{equation}
\label{eq:d-split}
d \;=\; d_S \,+\, d_H \,+\, d_W \,+\, d_D .
\end{equation}
This allocation reserves a dedicated subspace for each axis, so that slot and spatial phases can be rotated independently.

For each axis $a \in \{S,H,W,D\}$, sinusoidal frequencies are precomputed as
\begin{equation}
\label{eq:rope-freq}
\mathrm{freqs}_a(i) \;=\; \theta_a^{-\,2i/d_a},
\qquad i = 0,1,\ldots,\tfrac{d_a}{2}-1 ,
\end{equation}
where $\theta_a$ is a base period hyperparameter for axis $a$.
Following standard RoPE parameterization, larger $\theta_a$ values yield longer-wavelength components.
In all experiments, we simply set $\theta_S = N_{\text{slots}}$ and $\theta_H = H,\;\theta_W = W,\;\theta_D = D$, so that the lowest frequencies roughly span the full extent of each axis while higher indices encode finer spatial detail.

The 4D rotary embedding for a token at coordinates $(S,H,W,D)$ concatenates axis-wise embeddings:
\begin{equation}
\label{eq:rope-4d}
\begin{aligned}
\mathrm{RoPE}_{4D}(S,H,W,D)
&= \big[\, \mathrm{RoPE}_S(S), \, \mathrm{RoPE}_H(H), \\
&\quad\;\; \mathrm{RoPE}_W(W), \, \mathrm{RoPE}_D(D) \,\big].
\end{aligned}
\end{equation}
During self-attention, queries and keys are rotated as
\begin{equation}
\label{eq:rope-apply}
\begin{aligned}
Q' \;&=\; \mathrm{RoPE}_{4D}(S,H,W,D)\!\circ Q, \\
K' \;&=\; \mathrm{RoPE}_{4D}(S,H,W,D)\!\circ K,
\end{aligned}
\end{equation}
where $\circ$ denotes element-wise complex rotation on paired channels, and the rotated $(Q',K')$ are then used in standard dot-product attention.
The slot axis supplies a dedicated rotary phase per modality (e.g., $S{=}0$ for dose and $S{\ge}1$ for different conditions), while the spatial axes $(H,W,D)$ are shared across slots.
This design preserves a unified full-attention pass, but encourages structured cross-slot interactions (dose $\leftrightarrow$ CT/structures/beams) without adding extra parameters or attention blocks.

\subsection{$v$-parameterized Diffusion Objective}
\label{sec:method:diff}
Instead of predicting noise $\varepsilon$, we adopt the \emph{$v$-parameterization}, which provides a better balance of signal-to-noise across timesteps.
Given the forward process in \eqref{eq:forward}, we define
\begin{equation}
v(x_0,\varepsilon,t) \;=\; \alpha_t\,\varepsilon \;-\; \sigma_t\,x_0.
\label{eq:v-def}
\end{equation}
Given $(x_t^{(\tau)}, C, t)$ the model outputs $v_\theta(x_t^{(\tau)}, C, t)\in\mathbb{R}^{H \times W \times D}$ and is trained by
\begin{equation}
\label{eq:ldiff}
\begin{aligned}
\mathcal{L}_{\text{diff}}
~=~ \mathbb{E}_{t,\varepsilon,\tau,S}
\Big[\,
\big\|\,v_\theta(x_t^{(\tau)}, C, t) - v(x_0^{(\tau)},\varepsilon,t)\,\big\|_2^2
\Big].
\end{aligned}
\end{equation}
This choice is notationally convenient and improves optimization stability on 3D dose grids.
In particular, from $(x_t^{(\tau)}, v, t)$ we can recover $x_0$ and $\varepsilon$ exactly:
\begin{equation}
\label{eq:recovery}
x_0 = \alpha_t x_t - \sigma_t v, \qquad
\varepsilon = \sigma_t x_t + \alpha_t v.
\end{equation}

Thus $v_\theta$ still parameterizes the same forward process as noise prediction, but yields better-conditioned gradients across diffusion timesteps in practice.

At each training step we sample a target modality $\tau$ uniformly from $M$ and then draw a conditioning set $S \subseteq M \setminus \{\tau\}$ according to a simple curriculum on the number of observed modalities.
This ensures that any modality can act either as a target or as a condition, exposes the model to diverse conditioning patterns, and strengthens robustness to missing or incomplete inputs at inference time.

\subsection{Scorecard-aligned RL Post-training}
\label{sec:method:scardnft}

Pure diffusion training does not explicitly optimize for clinical objectives, such as precise PTV coverage or OAR sparing.
To bridge this gap, we propose \textbf{ScardNFT}, an RL-based post-training approach inspired by DiffusionNFT~\cite{Zheng2025DiffusionNFT}, which aligns generated dose distributions with clinical guidelines via differentiable, Scorecard-based rewards.

\nbf{Scorecard Reward}
We define a scalar clinical reward $r^{\text{raw}}$ from standardized plan-quality metrics.
For each anatomical structure $s \in \mathcal{S}$, the Scorecard specifies one of three metric types: \texttt{DoseAtVolume}, \texttt{VolumeAtDose}, or \texttt{MeanDose}, and maps each measured metric value into a normalized structure-specific score via a piecewise-linear function $\mathrm{score}_s(\cdot)$.
Let $\phi_s(y;C)$ denote the corresponding DVH-style statistic computed from a candidate dose $y$ under conditions $C$, and let $w_s \ge 0$ be a tunable importance weight.
The aggregate reward is then computed as a weighted sum:
\begin{equation}
\label{eq:raw-reward}
r^{\text{raw}}(y,C) ~=~ \sum_{s\in\mathcal{S}} w_s\,\mathrm{score}_s\!\big(\phi_s(y;C)\big),
\qquad y \equiv x_0^{(\texttt{dose})}.
\end{equation}
We adjust these clinical Scorecards using established radiotherapy templates for head and neck~\cite{varian2024han} and lung~\cite{varian2024lungconv}, covering critical dose--volume histogram (DVH) points, mean dose, and ring constraints.
When patient prescriptions differ from the scorecard template, we proportionally rescale PTV thresholds to the prescribed dose to maintain consistent clinical scoring criteria across cases.

\nbf{Normalization \& Anchors}
Plan quality can vary widely across patients and sites, so raw rewards must be normalized for stable learning.
Within each case, we first standardize per-structure scores and aggregate them into $r^{\text{raw}}$, and optionally apply site-wise normalization to account for systematic site differences.
To prevent reward hacking, we introduce two anchor terms: (i) a hinge penalty enforcing strict adherence to hard clinical constraints (e.g., minimum PTV D95 or maximum OAR thresholds), and (ii) an MAE anchor relative to available reference doses, discouraging trivial reductions in overall dose magnitude.
The resulting optimality probability $r$ used for RL is clipped to $[0,1]$:
\begin{equation}
\label{eq:optprob}
r \;=\; \tfrac{1}{2} + \tfrac{1}{2}\,
\mathrm{clip}\!\left(
\frac{\,r^{\text{raw}} - \mathbb{E}_{y\sim\pi_{\text{old}}}\,[r^{\text{raw}}]\,}{Z_C}\,,\, -1,\, 1\right),
\end{equation}
where $\pi_{\text{old}}$ denotes the current diffusion policy and $Z_C$ is a running estimate of reward dispersion under conditions $C$.
Thus $r \in [0,1]$ behaves as a Bernoulli-style optimality probability, with $r \approx 1$ indicating clinically preferred plans and $r \approx 0$ indicating poor plans under the same inputs.

\nbf{Policy Update \& Final Objective}
Starting from a pretrained diffusion checkpoint, we perform clinical preference-aligned policy updates via ScardNFT.
For each training case, we first draw $K$ candidate samples multiple independent initial noises using a deterministic ODE sampler (Flow/DPM-solver family, with scheduler state snapshots for reproducibility).
Each candidate dose $y$ is evaluated by computing its reward $r^{\text{raw}}(y,C)$, which is then transformed into an optimality probability $r$ via \eqref{eq:optprob}.
Inspired by DiffusionNFT \cite{Zheng2025DiffusionNFT}, we introduce two \emph{implicit} policy targets:
\begin{equation}
\small
\label{eq:implicit}
\tilde v_\theta^{+} \;=\; (1{-}\beta)\,v_{\text{old}} \;+\; \beta\,v_\theta,
\qquad
\tilde v_\theta^{-} \;=\; (1{+}\beta)\,v_{\text{old}} \;-\; \beta\,v_\theta,
\end{equation}
where $v_{\text{old}}$ is derived from the current model (with gradients stopped), $v_\theta$ is the new prediction, and $\beta \in (0,1]$ is a small mixing coefficient controlling how aggressively we move away from the old policy.
We then optimize a dual loss that increases likelihood for higher-rewarded samples and penalizes lower-rewarded ones:
\begin{equation}
\label{eq:nft}
\mathcal{L}_{\text{NFT}} ~=~ \mathbb{E}\Big[
\, r\,\|\tilde v_\theta^{+}- v\|_2^2 + (1{-}r)\,\|\tilde v_\theta^{-}- v\|_2^2 \Big],
\end{equation}
where $v$ denotes the ground-truth target from Eq.~\eqref{eq:v-def}.
The final training objective balances voxel-level diffusion consistency and clinical preference alignment:
\begin{equation}
\label{eq:final}
\mathcal{L}(\theta) ~=~ \mathcal{L}_{\text{NFT}}(\theta) \;+\; \lambda\,\mathcal{L}_{\text{diff}}(\theta),
\end{equation}
with $\lambda > 0$ controlling the strength of the RL-style update.

\nbf{Why this works}
Intuitively, $\mathcal{L}_{\text{diff}}$ preserves voxelwise fidelity to historical plans, while $\mathcal{L}_{\text{NFT}}$ reshapes the conditional diffusion score field, assigning higher likelihood to dose configurations that satisfy clinical Scorecards under identical conditions $C$.
The $v$-parameterization ensures well-scaled gradients across timesteps $(\alpha_t,\sigma_t)$, which we find empirically stabilizes ScardNFT updates, especially in high signal-to-noise ratio regimes.

\section{Experiments}
\label{sec:exp}

\begin{table*}[t]
\centering
\caption{Main results on GDP-HMM (validation \& test). Metrics: MAE (Gy; $\downarrow$), clinical Scorecard (Score) ($\uparrow$), PSNR (dB; $\uparrow$), SSIM ($\uparrow$), and LPIPS ($\downarrow$). The main table spans both columns; bold indicates the best in its block.}
\label{tab:main}
\setlength{\tabcolsep}{4pt}
\small
\resizebox{\textwidth}{!}{%
\begin{tabular}{l l c c c c c c c c c c c c c c}
\toprule
\multirow{2}{*}{Family} & \multirow{2}{*}{Method} & \multirow{2}{*}{Backbone} & \multirow{2}{*}{Pretrain} & \multirow{2}{*}{Any2Any} & \multirow{2}{*}{RL} & \multicolumn{5}{c}{Validation} & \multicolumn{5}{c}{Test}\\ 
\cmidrule(lr){7-11}\cmidrule(lr){12-16} 
& & & & & & MAE$\downarrow$ & Score$\uparrow$ & PSNR$\uparrow$ & SSIM$\uparrow$ & LPIPS$\downarrow$ & MAE$\downarrow$ & Score$\uparrow$ & PSNR$\uparrow$ & SSIM$\uparrow$ & LPIPS$\downarrow$ \\ 
\midrule 
Regression & Yasin (AAPM '25 Chal.) & MedNeXt & -- & -- & -- & 2.03 & 134.26 & 31.50 & 0.972 & 0.037 & 2.07 & 134.81 & 32.06 & 0.974 & 0.033 \\ 
Regression & tyxiong123 (AAPM '25 Chal.)& MedNeXt & -- & -- & -- & 2.18 & 133.95 & 29.89 & 0.929 & 0.067 & 2.20 & 134.10 & 30.62 & 0.925 & 0.064 \\ 
Regression & rcgao (AAPM '25 Chal.) & MedNeXt & -- & -- & -- & 2.05 & 133.40 & 31.24 & 0.972 & 0.051 & 2.08 & 133.62 & 31.77 & 0.972 & 0.048 \\ 
Regression & PVmed (AAPM '25 Chal.) & MedNeXt & -- & -- & -- & 2.15 & 133.78 & 31.40 & 0.971 & 0.034 & 2.17 & 133.99 & 31.53 & 0.970 & 0.031 \\ 
Regression & SKLSDE-BH (AAPM '25 Chal.) & nnUnet & -- & -- & -- & 2.25 & 133.02 & 31.40 & 0.972 & 0.040 & 2.27 & 133.20 & 31.34 & 0.968 & 0.039 \\ 
Regression & MedVision (AAPM '25 Chal.) & LDM & -- & -- & -- & 2.19 & 134.50 & 31.54 & 0.970 & 0.029 & 2.21 & 134.72 & 31.66 & 0.973 & 0.030 \\ 
\midrule 
Diffusion & MAISI + Ours & LDM & \checkmark & -- & -- & \textbf{1.89} & 135.89 & 32.02 & 0.975 & 0.028 & 1.95 & 135.23 & 32.13 & 0.976 & 0.026 \\ 
Diffusion & Ours (Conditional) & DiT & \checkmark & -- & -- & 2.07 & 135.41 & 31.88 & 0.973 & 0.031 & 2.12 & 134.60 & 32.01 & 0.974 & 0.029 \\ 
Diffusion & Ours (Any2Any) & DiT & \checkmark & \checkmark & -- & 1.90 & 136.22 & 32.43 & 0.977 & 0.025 & \textbf{1.93} & 135.36 & 32.60 & 0.978 & 0.023 \\ 
Diffusion & Ours (Any2Any+NFT) & DiT & \checkmark & \checkmark & \checkmark & 1.91 & \textbf{138.17} & \textbf{32.55} & \textbf{0.979} & \textbf{0.022} & \textbf{1.93} & \textbf{137.55} & \textbf{32.73} & \textbf{0.980} & \textbf{0.020} \\ 
\bottomrule
\vspace{-3em}
\end{tabular}}
\end{table*}

\subsection{Setup: Datasets, Splits, Metrics, and Training}
\label{sec:exp:setup}

We evaluate on the complete GDP-HMM Grand Challenge dataset \cite{AAPM2025GDPHMM}, comprising official \emph{training} (2,878 plans), \emph{validation} (356 plans), and \emph{test} (498 plans) splits for head-and-neck (HaN) and lung cancer sites, and the REQUITE dataset, comprising \emph{training} (5,100 plans) and \emph{test} (256 plans) splits for prostate cancer patients from \cite{seibold2019requite}, with mask-augmented plans re-optimized using the Eclipse Script API. Unless explicitly stated otherwise, voxelwise evaluations are conducted within the body mask.

\nbf{Preprocessing}
All data preprocessing steps adhere strictly to the official challenge geometry and voxel spacing. CT images undergo intensity clipping to [-1000, 1000] HU, patient-wise z-score normalization, and mask binarization. Beam and angle plates are rasterized onto the same grid. Before patch embedding, each modality is min-max scaled to [-1,1]. Comprehensive details regarding resampling, cropping, and normalization are provided in the supplementary material. The angle beam plates are created following \cite{Gao2023FlexibleCmGAN}, consistent with the GDP-HMM challenge.

\nbf{Primary metrics}
We evaluate performance using the following metrics: (i) voxelwise mean absolute error (MAE), computed within the body mask with a 5 Gy threshold following the challenge protocol \cite{gao2025automating}; (ii) clinically-informed plan quality Scorecards \cite{varian2024han,varian2024lungconv}, integrating key PTV and OAR metrics into a single scalar score; and (iii) standard image quality metrics including PSNR, SSIM, LPIPS, Dice, and 2D slice-level FID.

\nbf{Training details}
We utilize the Any2Any DiT architecture described in Sec.~\ref{sec:method:arch}. The backbone is Wan~2.1 (1.3B parameters), paired with VAE~2.1. Training proceeds through three distinct stages: (A) Any2Any pretraining with uniform target sampling and curriculum masking, (B) dose-only fine-tuning, and (C) ScardNFT post-training, balancing losses $\mathcal{L}_{\text{NFT}}$ and $\mathcal{L}_{\text{diff}}$ via a tunable hyperparameter $\lambda$. Further training details are provided in supplementary \ref{app:train}.

\subsection{Main Results on GDP-HMM Dataset}
\label{sec:exp:main}

The GDP-HMM benchmark contains a broad set of strong regression-based challenge entries built on MedNeXt, nnUNet, and LDM backbones (Table~\ref{tab:main}). These models represent the best supervised pipelines available but remain constrained by voxel-level training. To form a complete comparison spectrum, we additionally include (i) a conditional diffusion U-Net adapted from MAISI, and (ii) our own conditional DiT baseline that simply concatenates all conditioning modalities with dose for prediction.

Our Any2Any design yields an improvement in performance over both regression-style models and the concatenation-based diffusion baseline. Beyond numerical gains, this indicates that (1) jointly modeling all modalities in a unified diffusion space and (2) separating “role” (target vs.\ condition) through explicit embeddings are both essential for robust cross-modal dependency learning. Adding ScardNFT introduces consistent improvements in clinical alignment without degrading voxel-level fidelity, confirming that RL-guided updates reshape preference behavior rather than the underlying reconstruction quality.

\nbf{Single-step prediction analysis}
To better understand diffusion behavior in dose prediction, we compare single-step variants of x-pred and v-pred (Table~\ref{tab:any2any_simple}) and their scaling trends in Figure~\ref{fig:mae_line_charts}. While x-pred directly regresses $x_0$, the v-pred approach achieves substantially better performance, indicating a higher potential accuracy ceiling. Furthermore, iterative refinement from 1 to 10 steps consistently improves predictions, confirming that multi-step refinement remains essential for peak dosimetric performance.

\begin{table}[t]
\centering
\caption{Any2Any prediction under different prediction types and sampling steps.}
\label{tab:any2any_simple}
\setlength{\tabcolsep}{6pt}
\footnotesize
\begin{tabular}{lcccc}
\toprule
Method & Pred Type & Steps & MAE$\downarrow$ & Score$\uparrow$ \\
\midrule
Ours (Any2Any) & x-pred & 1  & 2.45 & 117.64 \\
Ours (Any2Any) & v-pred & 1  & 2.12 & 133.59 \\
\midrule
Ours (Any2Any) & v-pred & \textbf{10} & \textbf{1.91} & \textbf{138.17} \\
\bottomrule
\vspace{-1.5em}
\end{tabular}
\end{table}

\subsection{Results on REQUITE Prostate}
\label{sec:exp:prostate}

\begin{figure}[!t]
  \centering
 \includegraphics[width=\columnwidth]{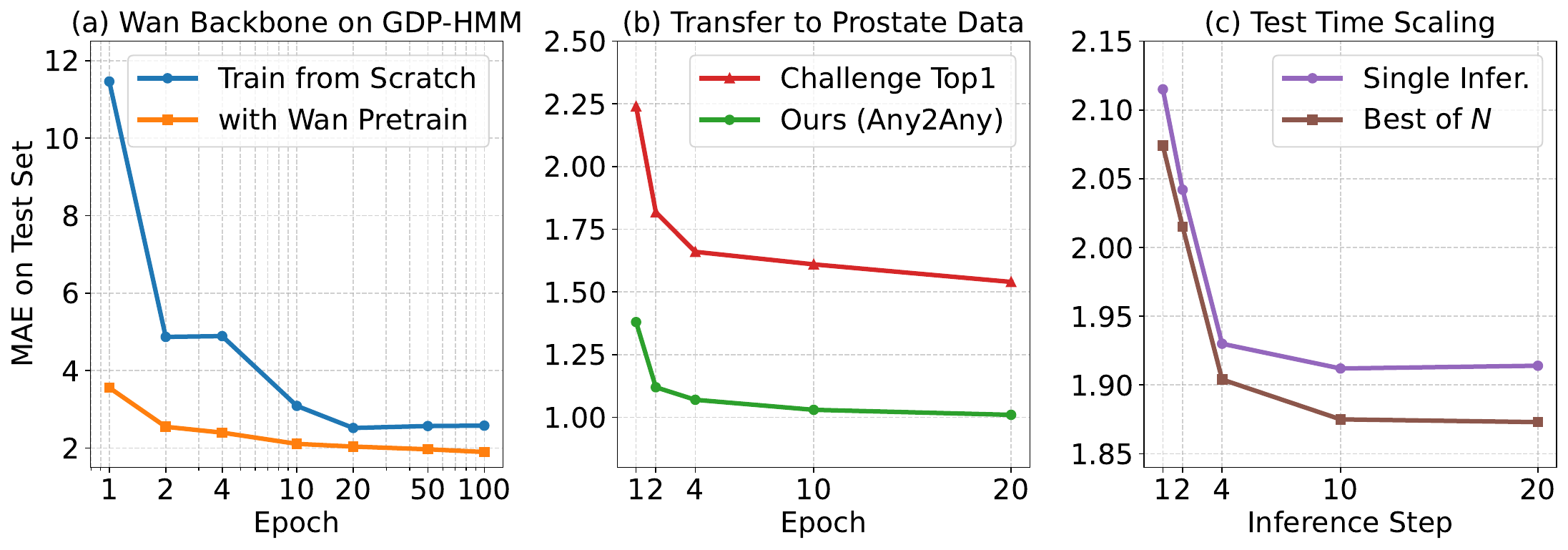}
  \vspace{-0.2in}
  \caption{MAE vs.\ training epochs / inference steps. The single figure contains three subplots: (left) pretrain vs.\ from-scratch across epochs, (middle) model-transfer finetuning curve, and (right) test-time scaling (single vs.\ best-of-$n$).}
  \label{fig:mae_line_charts}
\end{figure}

We further assess the knowledge transfer capabilities of our model by fine-tuning checkpoints pretrained on GDP–HMM (head-and-neck and lung) directly on the REQUITE prostate dataset. As shown in Table~\ref{tab:requite_prostate} and Figure~\ref{fig:mae_line_charts}(b), our \emph{Any2Any} diffusion model rapidly converges to superior performance compared to top regression baselines, achieving better accuracy in fewer epochs. This demonstrates our framework’s efficiency in leveraging pretrained representations to quickly adapt and achieve higher-quality predictions on new cancer sites. Moreover, adopting a best-of-$n$ sampling strategy provides additional gains, underscoring the model’s potential for further improvements through stochastic decoding.

\begin{table}[t]
\centering
\caption{Comparisons on REQUITE-Prostate. We report MAE (Gy; $\downarrow$), PSNR (dB; $\uparrow$), SSIM ($\uparrow$), and LPIPS ($\downarrow$). Both ours and baselines are pretrained with GDP-HMM and fine-tuned on prostate data. \dag\ denotes best-of-$n$.}
\label{tab:requite_prostate}
\setlength{\tabcolsep}{6pt}
\small
\resizebox{\linewidth}{!}{%
\begin{tabular}{lcccc}
\toprule
Method & MAE (Gy)$\downarrow$ & PSNR (dB)$\uparrow$ & SSIM$\uparrow$ & LPIPS$\downarrow$ \\
\midrule
Yasin (AAPM '25 Chal.)      & 1.48 & 34.53 & 0.956 & 0.026 \\
tyxiong123 (AAPM '25 Chal.) & 1.37 & 34.74 & 0.957 & 0.023 \\
rcgao (AAPM '25 Chal.)      & 1.53 & 33.51 & 0.953 & 0.031 \\
PVmed (AAPM '25 Chal.)      & 1.52 & 33.54 & 0.950 & 0.030 \\
SKLSDE-BH (AAPM '25 Chal.)  & 1.65 & 32.74 & 0.944 & 0.034 \\
MedVision (AAPM '25 Chal.)  & 1.48 & 34.42 & 0.949 & 0.032 \\
\midrule
\textbf{Ours (Any2Any)}               & \textbf{1.01} & \textbf{36.80} & \textbf{0.963} & \textbf{0.012} \\
\textbf{Ours (Any2Any)}$^{\dag}$       & \textbf{0.97} & \textbf{37.09} & \textbf{0.965} & \textbf{0.011} \\
\bottomrule
\vspace{-1.5em}
\end{tabular}}
\end{table}

\begin{table}[t]
\centering
\caption{Component ablations on validation set}
\label{tab:abl_components}
\setlength{\tabcolsep}{4pt}
\small
\resizebox{\linewidth}{!}{%
\begin{tabular}{lccccccc}
\toprule
Variant 
& Pre. 
& Any2Any
& RoleEmb
& FullAttn
& PatchEmb
& 4D RoPE
& MAE$\downarrow$/Score$\uparrow$ \\
\midrule
From scratch (baseline)     
& --         & --         & -- & -- & -- & --         
& 2.58 / 119.82 \\
+ Pretrain                  
& \checkmark & --         & -- & -- & -- & --         
& 2.07 / 135.41 \\
+ Any2Any 
& \checkmark & \checkmark & \checkmark & \checkmark & \checkmark & \checkmark 
& 1.90 / 136.22 \\

- Role Emb                  
& \checkmark & \checkmark & --         & \checkmark & \checkmark & --         
& 2.01 / 134.04 \\
- Full Attention            
& \checkmark & \checkmark & \checkmark & --         & \checkmark & --         
& 2.15 / 130.80 \\
- Patch Embed               
& \checkmark & \checkmark & \checkmark & \checkmark & --         & --         
& 2.02 / 134.71 \\
- 4D RoPE                   
& \checkmark & \checkmark & \checkmark & \checkmark & \checkmark & --       
& 1.96 / 135.27 \\
\midrule
\textbf{Ours+ScardNFT} 
& \checkmark & \checkmark & \checkmark & \checkmark & \checkmark & \checkmark 
& \textbf{1.91 / 138.17} \\
\bottomrule
\vspace{-2em}
\end{tabular}}
\end{table}

\begin{table}[t]
\centering
\setlength{\tabcolsep}{6pt}
\caption{
Single-modality prediction under the remaining-1 (predict-one) setting.
\emph{What this table shows:} each modality is predicted from all the others.
CT uses FID; segmentation-like modalities use Dice; Dose and Beam Plate use MAE only.
}
\label{tab:remain1}
\scriptsize
\begin{tabular}{lcccc}
\toprule
Method & Output Modality & FID$\downarrow$ & Dice$\uparrow$ & MAE$\downarrow$ \\
\midrule
Ours (Any2Any) & CT          & 1.47 & --    & --    \\
Ours (Any2Any) & PTV mask    & --   & 72.13 & --    \\
Ours (Any2Any) & OAR mask    & --   & 54.63 & --    \\
Ours (Any2Any) & Body mask   & --   & 95.09 & --    \\
Ours (Any2Any) & Dose        & --   & --    & 1.91  \\
Ours (Any2Any) & Beam Plate  & --   & --    & 1.54  \\
\bottomrule
\vspace{-2.0em}
\end{tabular}
\end{table}

\begin{figure}[!t]
  \centering
  \vspace{-0.1in}
  \includegraphics[width=0.88\columnwidth]{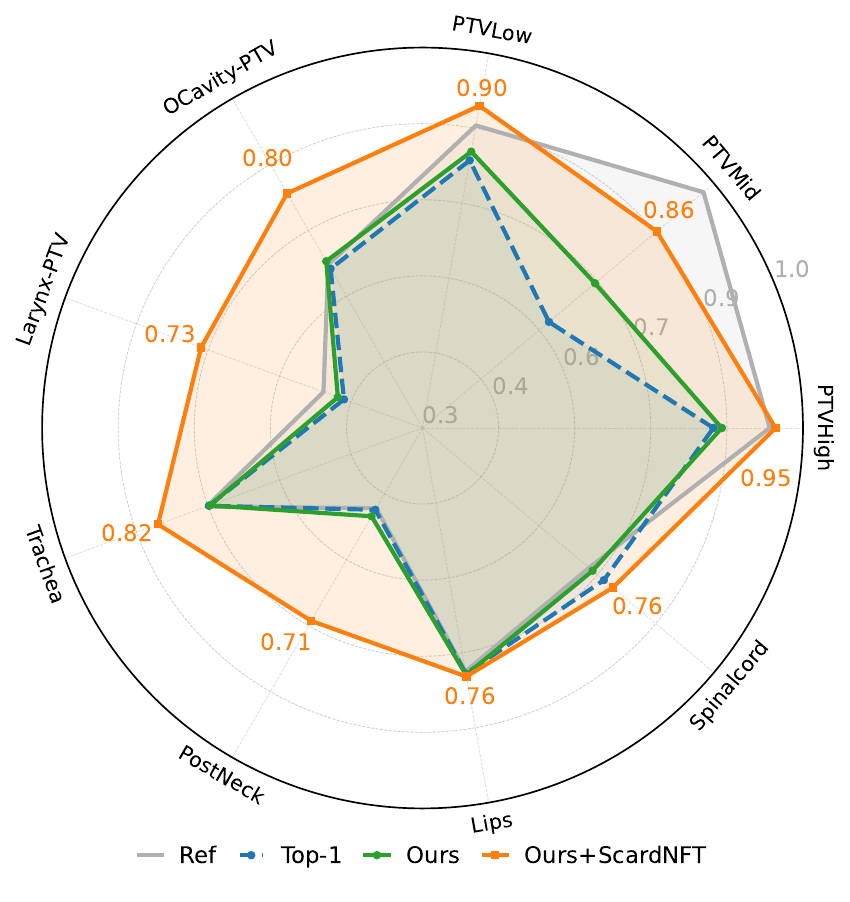}
  \vspace{-0.2in}
  \caption{
  Per-structure scorecard value comparison of head-and-neck plans. 
  The plot contrasts reference, the challenge Top-1 baseline, our diffusion model,
  and our RL-enhanced variant (Ours+ScardNFT), showing how reinforcement learning can
  improve alignment with institutional planning objectives.
  }
  \label{fig:scorecard_spider}
  \vspace{-1.0em}
\end{figure}
\subsection{Ablations and Supporting Studies}
\label{sec:exp:abl}

Detailed ablation studies in Table~\ref{tab:abl_components} and Figure~\ref{fig:mae_line_charts}(a) clarify the contributions of key components and design choices. Introducing pretraining substantially accelerates convergence compared to training from scratch and achieves significantly better final performance, confirming the value of leveraging pretrained knowledge. Transitioning from conditional diffusion to the unified Any2Any training paradigm further boosts performance, demonstrating the effectiveness of flexible cross-modality modeling. Removing critical components notably reduces performance. Omitting the role embeddings, which explicitly indicate the target and conditioning roles, clearly degrades voxelwise accuracy and clinical scores. Similarly, replacing full attention with causal attention weakens the model's ability to capture cross-modal dependencies. Performance also declines without modality-specific patch embeddings, emphasizing their importance in preserving detailed input modality information. The proposed 4D RoPE positional embedding further boosts performance by uniquely identifying each condition in both spatial and temporal dimensions. Although 4D RoPE appears related to role embeddings, it specifically distinguishes among different input modalities spatially, whereas role embeddings explicitly inform the model of modalities serving as either conditions or targets. Finally, integrating ScardNFT post-training improves clinical alignment, significantly enhancing clinical preference scores without compromising MAE.

Figure~\ref{fig:mae_line_charts} further elucidates our model’s training efficiency and transferability. Figure~\ref{fig:mae_line_charts}(a) clearly illustrates the substantial efficiency gains from Wan pretraining compared to training from scratch, reaching lower MAE values with significantly fewer epochs. Figure~\ref{fig:mae_line_charts}(b) demonstrates the model’s strong adaptability when transferring to the prostate dataset, rapidly converging and consistently outperforming the top GDP–HMM challenge solution. Finally, Figure~\ref{fig:mae_line_charts}(c) highlights the benefit of a best-of-$n$ inference strategy, providing a clear accuracy improvement over single inference.

As illustrated in Figure~\ref{fig:scorecard_spider}, ScardNFT post-training notably enhances clinical alignment across key anatomical structures. Compared to our baseline without ScardNFT and the top regression method, the ScardNFT variant achieves consistently higher clinical scores for PTV coverage and OAR sparing, while maintaining identical voxelwise MAE performance (Table~\ref{tab:main}). This improvement significantly benefits downstream clinical tasks such as plan optimization and automated quality assurance.

\begin{figure}[!t]
  \centering
  \includegraphics[width=0.97\columnwidth]{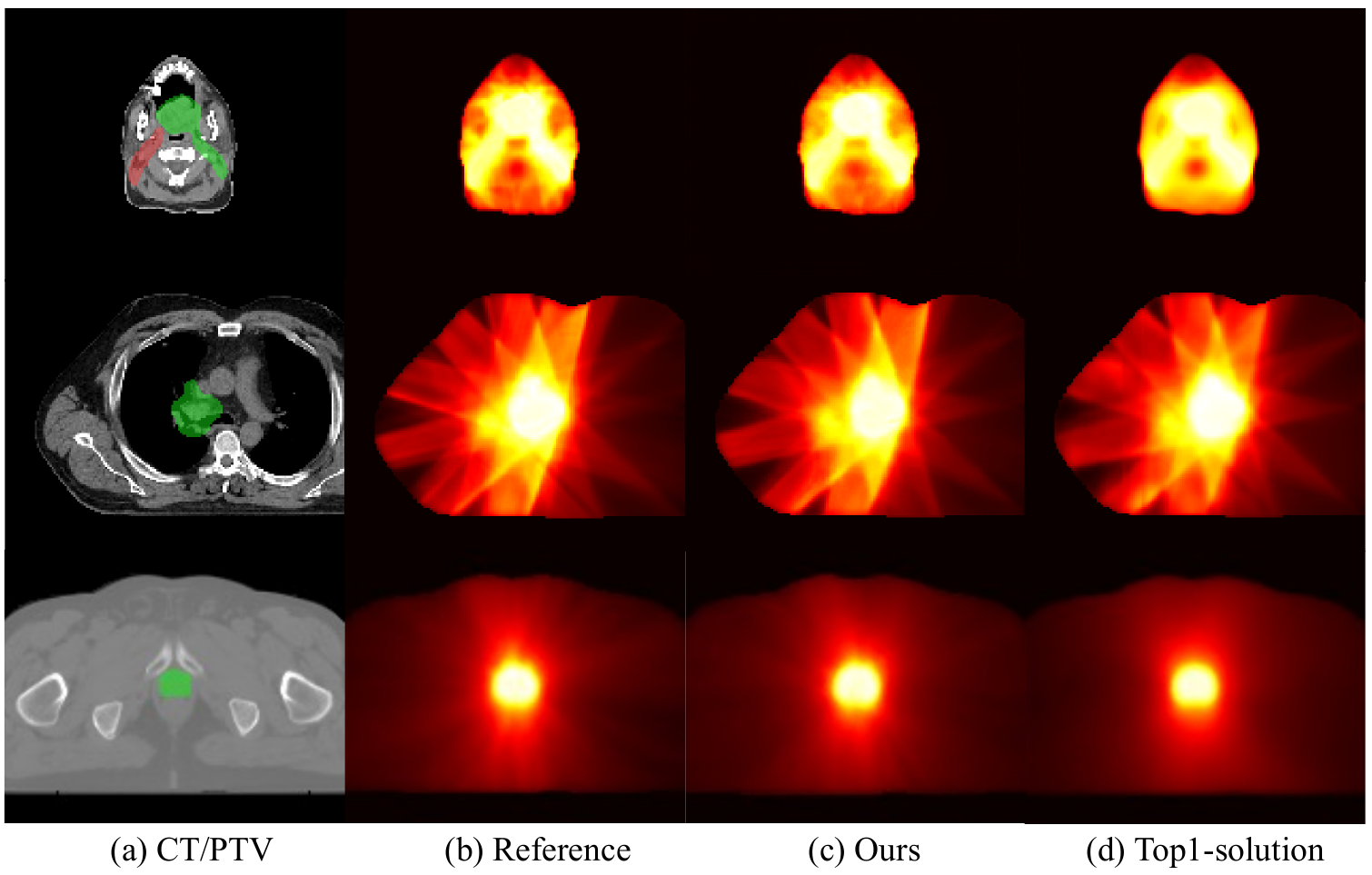}

  \caption{
    Qualitative predictions on GDP-HMM and REQUITE (top: head-and-neck, middle: lung, bottom: prostate). Columns: Input CT, reference dose, our Any2Any+NFT, Top-1 baseline. Our method achieves superior target conformity.
  }
  \vspace{-1.5em}
  \label{fig:qualitative_gdphmm}
\end{figure}

Qualitative examples in Figure~\ref{fig:qualitative_gdphmm} further confirm our method's robustness. Compared to the leading regression baseline, Our Any2Any+ScardNFT model produces more clinically realistic dose distributions, improving conformity around targets (head-and-neck), reducing artifacts (lung), and mitigating oversmoothing (head-and-neck and prostate). More visualizations are provided in Supplementary \ref{sec:more_visual}.

Table~\ref{tab:remain1} assesses our model's robustness under the remaining-1 prediction scenario, where each modality is predicted from all other modalities. Our model consistently generates high-quality predictions across imaging (CT), segmentation (PTV, OAR, Body mask), and dose-related modalities, validating its strong cross-modal generative capability.




\section{Conclusion and Discussion}
\nbf{Conclusion} We presented DiffKT3D, an Any2Any 3D diffusion framework for voxel-wise radiotherapy dose prediction and cross-modal imputation. By integrating pretrained diffusion priors through a modality-aware conditioning interface and aligning predictions to clinical guidelines via reinforcement learning post-training, DiffKT3D consistently outperformed strong regression and diffusion baselines on the GDP–HMM challenge and REQUITE datasets.

\nbf{Discussion} Beyond gains in dose prediction accuracy, inference efficiency, and preference compliance, our findings suggest broader methodological implications.

\textbf{\textit{Diffusion priors}} trained on cross-domain datasets can transfer effectively to specialized tasks, mitigating domain shift and reducing training cost. In our experiments, priors trained on CT (MAISI) or video (Wan) already improve dose generation. While feature-extraction foundation models such as the DINO family \cite{Oquab2023DINOv2,simeoni2025dinov3} demonstrate cross-domain robustness, 3D generative priors remain underexplored. Our results show that public diffusion backbones are effective initializations for RT dose modeling, avoiding the need to train large 3D models from scratch.

\textbf{\textit{RL post-training}} is widely used for LLMs and diffusion-based text-to-image alignment. Here, we apply it to clinical decision-making tasks, where rewards are derived directly from clinical protocols. Our scorecard-aligned reward translates guideline criteria into optimization signals and should extend to other medical tasks.

The unified \textbf{\textit{Any2Any}} conditional architecture provides a flexible paradigm for handling diverse multi-modal scenarios beyond RT dose prediction, underscoring the potential of our approach as a generalizable framework for conditional generative modeling.

\nbf{Limitation and Future Work}
One limitation of DiffKT3D is its computational efficiency. The Wan 1.3B backbone with full 3D attention is inherently expensive, and even with a 4-step sampler, end-to-end inference for a full 3D dose takes about 10 s on a single GPU. While this remains far faster than optimization-based systems (e.g., 15–30 min for head-and-neck VMAT), future work will investigate efficiency-oriented strategies such as lighter backbones, structured or sparse attention, token pruning, and distilling the Any2Any DiT into compact student models.

The training objective can be further enhanced by incorporating dose-specific loss functions, such as DVH-based terms \cite{jhanwar2022domain,Gao2023FlexibleCmGAN,nguyen2020incorporating} and weighted MAE \cite{gao2025automating}, into our diffusion training pipeline. Additionally, validating the model’s effectiveness in real clinical settings remains an important direction for future work. Finally, the Any2Any design of DiffKT3D extends beyond dose prediction; applying it to other stages of the radiotherapy planning pipeline (e.g., leaf sequencing) and to broader generative tasks represents a natural and promising next step.

\textbf{Disclaimer.} The information in this paper is based on research results that are not commercially available. Future commercial availability cannot be guaranteed. 

\textbf{Acknowledgement}: We thank all the contributors to the REQUITE project, including the patients, clinicians and nurses. The core REQUITE consortium consists of David Azria, Erik Briers, Jenny Chang-Claude, Alison M. Dunning, Rebecca M. Elliott, Corinne Faivre-Finn, Sara Gutiérrez-Enríquez, Kerstie Johnson, Zoe Lingard, Tiziana Rancati, Tim Rattay, Barry S. Rosenstein, Dirk De Ruysscher, Petra Seibold, Elena Sperk, R. Paul Symonds, Hilary Stobart, Christopher Talbot, Ana Vega, Liv Veldeman, Tim Ward, Adam Webb and Catharine M.L. West.

{\small
\bibliographystyle{ieeenat_fullname}
\bibliography{11_references}
}

\ifarxiv \clearpage \onecolumn \appendix
\addtocontents{toc}{\protect\setcounter{tocdepth}{2}}
\renewcommand{\contentsname}{Supplementary Contents}
\vspace*{-2mm}
{\tableofcontents}
\vspace*{2mm}\hrule\vspace*{3mm}
\newpage
\section{Detailed Model Structures}
\label{app:struct}

As shown in Figure~\ref{fig:detail_model}, our DiffKT3D adopts a VAE--DiT hybrid architecture~\cite{kingma2014vae,peebles2023dit,wan2025wan}. For illustration we depict three volumetric inputs $X_a$, $X_b$, and $X_g$ corresponding to CT, structure masks (PTV and OARs), and dose, respectively; the same pipeline applies to all available modalities. Each volume is first passed through a frozen 3D VAE encoder~\cite{kingma2014vae} to obtain compact latent representations. These latent grids are then patchified into token sequences and concatenated before being fed into a stack of DiT blocks~\cite{peebles2023dit}. The diffusion process operates entirely in this latent-token space. After denoising, the output tokens are reshaped back into latent feature maps $V_a$, $V_b$, and $V_g$, which are decoded by the corresponding VAE decoders to recover volumetric predictions at the original spatial resolution.

The right panel of Figure~\ref{fig:detail_model} details a single DiT block. Each block follows a transformer-style design~\cite{vaswani2017attention} with self-attention and a feed-forward network (FFN), all wrapped by residual connections. In the original Wan~2.1 backbone~\cite{wan2025wan}, each block also contains a cross-attention layer that lets vision tokens attend to text tokens. DiffKT3D does not use any language or text conditioning, so we remove this cross-attention module and let all tokens from all volumetric modalities (CT, masks, dose, and other channels) jointly interact through the shared self-attention layers. A shared timestep--role embedding (encoding the diffusion step and whether a token is a target or a conditioning token) is processed by a small MLP to produce modulation vectors. These vectors drive FiLM-like layers~\cite{perez2018film}: we apply scale-and-shift operations to the normalized tokens before the self-attention and FFN modules, and \textit{Scale}-only gates on the residual outputs of self-attention and FFN. This modulation allows each DiT block to dynamically control feature amplification or suppression across timesteps and roles while keeping the overall architecture lightweight and stable to train.

\begin{figure}[h]
    \centering
    \includegraphics[width=\linewidth]{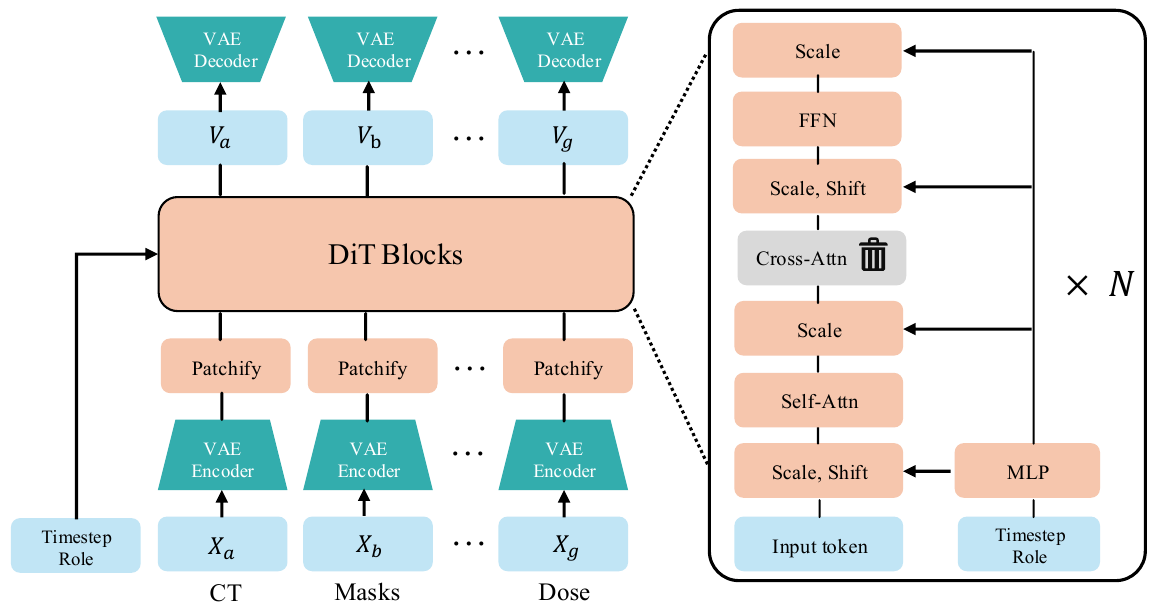}
    \caption{Architecture of the proposed VAE--DiT-based conditional diffusion model DiffKT3D. 
    Left: multi-branch VAE--DiT pipeline for CT ($X_a$), structure masks ($X_b$), and dose ($X_g$) with their corresponding latent outputs $\{V_a, V_b, V_g\}$. 
    Right: structure of a single DiT block with timestep--role modulation using FiLM-style (\textit{Scale, Shift}) layers and residual gates (\textit{Scale}). The Any2Any gating and noisy latent are not shown for simplification. We remove cross-attention layers from original Wan DiT blocks because DiffKT3D does not use language tokens.}
    \label{fig:detail_model}
\end{figure}

\newpage
\section{Training Details}
\label{app:train}
\subsection{Data Sources}

We train and evaluate DiffKT3D on the official GDP--HMM Grand Challenge dataset \cite{gao2025automating} for head-and-neck and lung cancer and on the REQUITE prostate cohort \cite{seibold2019requite}, strictly following the organizers’ definition of the voxel grid (spacing, orientation, and cropping box) and the body mask. The REQUITE cohort plans were re-optimized in Varian Eclipse ESAPI under multiple planning configurations, yielding multiple plans per patient. CT images have been clipped to \([-1000, 1000]\) HU, and normalized on a per-patient basis before loading into AI model; all structure masks, beam plates, and angle plates are rasterized onto the same grid as the reference dose. To obtain a fixed field-of-view compatible with the Wan~2.1 VAE, we crop a \(97 \times 128 \times 160\) 3D region of interest around the PTV isocenter for every case. The in-plane size \(128 \times 160\) matches the challenge bounding box, while the depth of \(97\) voxels is chosen as \(4d+1\) so that the downsampled latent depth satisfies the causal attention constraint in the Wan~2.1 VAE. After cropping, all modalities are linearly scaled to the range \([-1, 1]\) before being passed into the frozen VAE encoder, matching the expected input range of the pretrained backbone. 

\subsection{Conditioning Modalities and Structure Selection}

Each patient is represented by up to seven modalities (see the modality visualization in the appendix of \cite{gao2025automating}):
\begin{equation*}
\{\,\text{CT},~\text{PTV},~\text{OAR masks},~\text{body mask},~\text{dose},~\text{beam plate},~\text{angle plate}\,\}.
\end{equation*}
The ``PTV'' channel encodes the optimized planning target volumes after any site-specific post-processing. Beam and angle plates follow the official GDP--HMM implementation and provide beam geometry and gantry angle information on the same voxel grid as the dose.

To make supervision consistent across disease sites, we standardize the set of OARs used during training. As in the challenge data, we retain up to about 30 OARs for head-and-neck, and 7 OARs for lung plans. For prostate plans, we retain four OARs: bladder, rectum, femoral head (left), and femoral head (right). 
All masks are stored as floating-point channels and normalized jointly with the other modalities to \([-1, 1]\) before patch embedding.

\subsection{Backbone Adaptation and Supervised Training}

We initialize DiffKT3D from the public Wan~2.1 DiT+VAE checkpoint \cite{wan2025wan}, whose DiT backbone follows the scalable diffusion transformer design of Peebles and Xie \cite{peebles2023dit}, and keep the VAE completely frozen throughout all experiments. On top of Wan’s 3D patch embedding, we introduce seven modality-specific 3D patch-embedding heads, one per modality in the set above. Each head has the same architecture as the original Wan patch embed but uses separate parameters, mapping the latent grids (or their noised versions for target modalities) into tokens of hidden dimension \(D\).

To support Any2Any training, we augment the backbone with:
(i) a learnable binary role embedding that tags each token as either target or condition and is injected via the shared AdaLayerNorm modulator by adding it to the timestep embedding, and
(ii) a 4D RoPE positional encoding that assigns rotary phases along a slot axis (modality ID) and the three spatial axes \((H, W, D)\).
These additions are lightweight and leave the Wan DiT block structure unchanged; only the DiT blocks and the new embedding layers are fine-tuned on RT data.

\subsection{Baselines and Fairness Protocol}

In GDP--HMM challenge, regression baselines are the top challenge entries built on MedNeXt~\cite{roy2023mednext}, nnU-Net~\cite{isensee2021nnunet}, and latent diffusion backbones~\cite{rombach2022highresolution}, and we evaluate them using the official model weights released by the organizers. Diffusion baselines include an MAISI-based conditional U-Net~\cite{guo2025maisi} and a conditional DiT variant that concatenates all conditioning modalities with the dose channel~\cite{peebles2023dit}. All methods operate on exactly the same cropped \(97 \times 128 \times 160\) volumes and use the same set of modalities and OAR selection as DiffKT3D.

For the REQUITE prostate experiments, where no challenge leaderboard is available, we initialize all baselines from their GDP--HMM-trained checkpoints\footnote{\url{https://huggingface.co/Jungle15/GDP-HMM_baseline/tree/main/participants_solutions}} and fine-tune them on prostate data under the same schedule as our model: identical preprocessing, crop size, effective batch size, and number of epochs. Our internal regression and diffusion variants are also trained with the same protocol. This setup ensures that performance differences come from the model design (Any2Any conditioning, role embeddings, 4D RoPE, and post-training) rather than from data handling or compute budget.

\subsection{RL Post-training (ScardNFT)}
\label{app:sub_post_train}

After supervised training we perform a lightweight RL-style post-training stage using the ScardNFT objective, which instantiates the DiffusionNFT formulation~\cite{Zheng2025DiffusionNFT} on our clinical scorecard. For each patient, we generate candidate dose predictions with a 10-step deterministic sampler from the Flow-Matching/DPM-Solver family~\cite{lipman2022flowmatching,lu2022dpmsolver}, starting from multiple initial noises and evaluate each candidate using the clinically informed Scorecard together with a voxel-wise mean absolute error (MAE) anchor. Based on these scores, we construct four positive/negative sample pairs per case and optimize the DiffusionNFT-style loss described in the main text.

To keep this stage efficient and stable, we avoid full-parameter fine-tuning. Instead, we insert rank-64 LoRA adapters into all self-attention and feed-forward layers of the DiT backbone and update only these adapters together with the small modulation networks, keeping the original Wan weights frozen~\cite{hu2022lora}. This post-training is only a small fraction of overall training cost but yields improved clinical preference alignment reported in the main paper without sacrificing voxel-level accuracy.

\subsection{Optimization and Hyperparameters}

 The base DiffKT3D models (before ScardNFT post-training) are obtained from a single Flow-Matching training run with a $v$-prediction objective in the latent space. We train for 100 epochs on eight B200 GPUs with data parallelism, using a per-GPU batch size of 1 (effective batch size 8). The model is optimized with Adam ($\beta_1 = 0.9$, $\beta_2 = 0.999$, $\epsilon = 10^{-8}$, no weight decay) and a constant learning rate of $1\times 10^{-4}$ after a linear warm-up of 500 steps. We use 1{,}000 training timesteps with a flow-shift parameter of $3.0$, and enable timestep embedding with an additional $\sigma$-embedding. Training is performed in bfloat16 mixed precision with gradient accumulation disabled and gradient clipping at a global norm of 0.1. We do not use classifier-free guidance or dropout in the conditioning pathways (CFG dropout probability set to 0). The subsequent ScardNFT post-training stage uses the same optimizer but only updates the LoRA adapters and modulation MLPs described in Sec. \ref{app:sub_post_train}.

\newpage
\section{More Visual Results}
\label{sec:more_visual}


We provide additional qualitative examples in Figure~\ref{fig:supp_visual_qualitative} to complement the quantitative results in the main paper. For three representative patients from the head-and-neck, lung, and prostate cohorts, we display the CT slice with contoured PTV and OARs, followed by the ground-truth dose distribution, the prediction from DiffKT3D, and the prediction from the challenge top-1 baseline. Below each row we plot the DVHs of the target and selected OARs, and we also show voxel-wise absolute error maps between the predictions and the ground-truth dose. These examples illustrate that, across disease sites, DiffKT3D better preserves PTV coverage while reducing hot spots in nearby OARs and body compared with the top-1 baseline. 

For data sample visualization, you may refer to the work of challenge data preparation \cite{gao2025automating}.


\newpage
\begin{figure}[h]
    \centering
    \includegraphics[width=\linewidth]{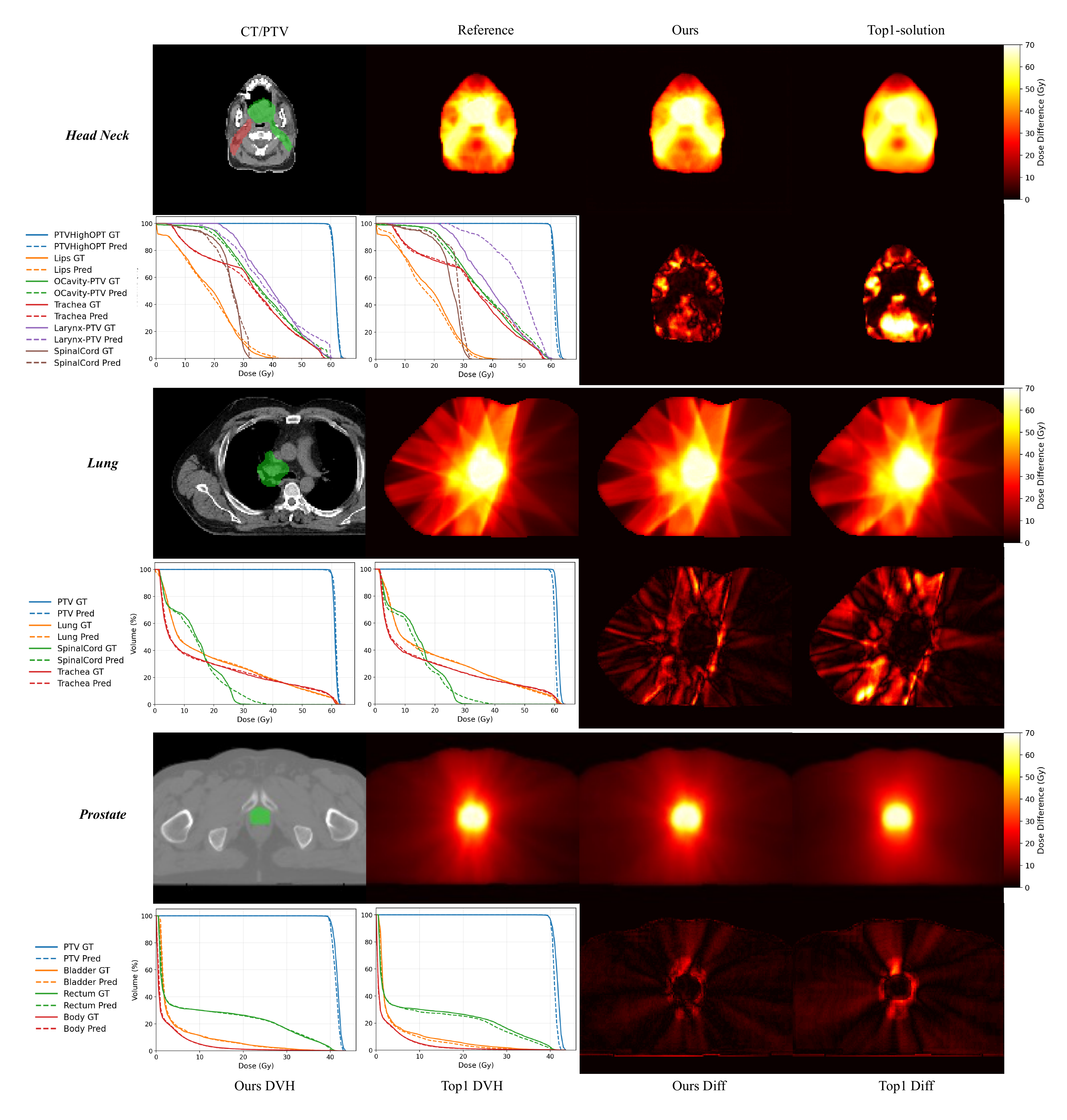}
    \caption{Qualitative comparison on representative head-and-neck, lung, and prostate cases. For each case we show CT with delineated PTV/OARs, ground-truth dose, DiffKT3D prediction, and the challenge top-1 baseline, together with DVHs and voxel-wise absolute error maps. Color bars are in Gy.}
    \label{fig:supp_visual_qualitative}
\end{figure}

\newpage
\section{Additional Baseline Comparisons}
\label{sec:additional_baselines}

To provide a broader assessment of DiffKT3D against alternative diffusion-based conditioning strategies, we evaluate three additional baselines on the GDP--HMM dataset: (i)~a \emph{3D ControlNet}~\cite{Zhang2023ControlNet} that applies ControlNet-style conditioning branches to the Wan~2.1 DiT backbone, (ii)~a \emph{2D slice-wise diffusion} model following prior RT dose prediction works~\cite{Feng2023DiffDP} that processes each axial slice independently with a 2D diffusion backbone, and (iii)~a \emph{LoRA-only} variant that replaces full DiT fine-tuning with rank-64 LoRA adapters under the same Any2Any paradigm. Table~\ref{tab:supp_extra} reports performance alongside the main models from the paper, together with per-case inference time and peak GPU memory on a single H100.

\begin{table}[h]
    \centering
    \caption{Extended comparison on GDP--HMM (validation set). All diffusion models use 10-step sampling. Inference time and peak GPU memory are measured per case on a single H100 GPU. Data loading time is included.}
    \label{tab:supp_extra}
    \setlength{\tabcolsep}{3pt}
    \small
    \begin{tabular}{lcccc}
        \toprule
        Method & MAE$\downarrow$ & Score$\uparrow$ & Time\,(s)$\downarrow$ & Mem\,(GB)$\downarrow$ \\
        \midrule
        Challenge Top-1 (regression) & 2.03 & 134.26 & 3.19 & 3.08 \\
        Ours (MAISI, CT pretrain) & 1.89 & 135.89 & 3.16 & 2.68 \\
        Ours (Conditional DiT) & 2.07 & 135.41 & 8.35 & 6.67 \\
        Ours (Any2Any, full fine-tuning) & 1.90 & 136.22 & 16.04 & 8.48 \\
        Ours (Any2Any + ScardNFT) & 1.91 & 138.17 & 16.42 & 8.70 \\
        \midrule
        3D ControlNet & 2.42 & 125.79 & 17.65 & 10.24 \\
        2D slice-wise diffusion & 2.14 & 132.90 & 17.95 & 32.40 \\
        LoRA on Wan DiT + Any2Any & 2.26 & 132.44 & 16.42 & 8.70 \\
        \bottomrule
    \end{tabular}
\end{table}

\paragraph{Analysis.}
The 3D ControlNet approach, while effective in natural image domains, performs substantially worse (MAE~2.42, Score~125.79) when applied to heterogeneous RT modalities. We attribute this to the fundamental mismatch between the ControlNet design---which assumes a single conditioning modality of the same domain as the generation target---and the RT setting where multiple structurally diverse modalities (CT, binary masks, beam plates, angle encodings) must jointly guide generation. The ControlNet copy-branch architecture cannot flexibly distinguish between these heterogeneous inputs, and its additional parameters increase memory overhead without compensating performance gains.

The 2D slice-wise diffusion baseline achieves reasonable voxelwise accuracy (MAE~2.14) but suffers from inter-slice inconsistency and significantly worse clinical Scorecard alignment (132.90). Processing each axial slice independently discards 3D spatial context that is crucial for dose conformality, particularly in complex head-and-neck geometries where dose gradients span many slices. Additionally, this approach requires substantially more GPU memory (32.40\,GB vs.\ 8.70\,GB for our full model) due to the need to process all slices sequentially and stitch results.

The LoRA-only variant (MAE~2.26, Score~132.44) demonstrates that parameter-efficient fine-tuning alone is insufficient to fully adapt the pretrained Wan backbone for the RT domain when used throughout the entire training pipeline. Full fine-tuning of the DiT blocks during the main supervised training stage remains essential for closing the large domain gap between natural video and medical dose data. However, as discussed in Section~\ref{app:sub_post_train}, LoRA proves effective for the lightweight ScardNFT post-training stage, where it stabilizes RL updates without degrading the well-trained base model.

\paragraph{Computational context.}
In a clinical RT planning workflow, optimization-based planning typically requires 5--30 minutes per case depending on complexity and beam arrangement. In this context, the inference time of DiffKT3D, even with 10-step sampling, can be well within practical deployment thresholds. Dose prediction may serve as an initialization or quality-assurance tool rather than the final deliverability, and faster single-step inference at second-level can be used for interactive exploration when speed is preferred over peak accuracy. We note that further runtime reductions are achievable through CUDA/C++ deployment optimization and model distillation, which we leave for future work.

\newpage
\section{Extended Ablation Studies}
\label{sec:extended_ablations}

We present two additional ablation studies that complement the analyses in the main paper: (i)~the effect of noise-prediction ($\epsilon$-prediction) parameterization, and (ii)~the impact of VAE adaptation strategies.

\subsection{Noise Prediction Parameterization}

Table~\ref{tab:any2any_simple} in the main paper compares $x_0$-prediction and $v$-prediction under the Any2Any framework. Here we additionally evaluate $\epsilon$-prediction (noise prediction), which is the most common parameterization in standard diffusion literature~\cite{ho2020denoising}, to provide a complete picture of prediction-type choices.

\begin{table}[h]
    \centering
    \caption{Comparison of prediction parameterizations on GDP--HMM (validation). Results for $x_0$-pred and $v$-pred are reproduced from Table~\ref{tab:any2any_simple} in the main paper for convenience.}
    \label{tab:supp_pred}
    \setlength{\tabcolsep}{6pt}
    \small
    \begin{tabular}{lccc}
        \toprule
        Prediction Type & Steps & MAE$\downarrow$ & Score$\uparrow$ \\
        \midrule
        $x_0$-pred & 1 & 2.45 & 117.64 \\
        $\epsilon$-pred & 1 & 2.16 & 133.09 \\
        $v$-pred & 1 & 2.12 & 133.59 \\
        \midrule
        $\epsilon$-pred & 10 & 1.93 & 137.82 \\
        $v$-pred & 10 & \textbf{1.91} & \textbf{138.17} \\
        \bottomrule
    \end{tabular}
\end{table}

\paragraph{Analysis.}
Both $\epsilon$-pred and $v$-pred substantially outperform $x_0$-pred in the single-step regime, confirming that direct signal regression is suboptimal for the flow-matching framework adopted by DiffKT3D. Between $\epsilon$-pred and $v$-pred, the gap is modest at 1~step (MAE: 2.16 vs.\ 2.12; Score: 133.09 vs.\ 133.59) and narrows further at 10~steps (MAE: 1.93 vs.\ 1.91; Score: 137.82 vs.\ 138.17). The consistent advantage of $v$-pred aligns with observations in the video generation literature~\cite{salimans2022progressive}, where $v$-parameterization improves training stability and sample quality under flow-matching objectives. We therefore adopt $v$-pred as the default for all main experiments.

\subsection{VAE Adaptation Strategies}

DiffKT3D keeps the pretrained Wan~2.1 VAE entirely frozen during training. Here we evaluate whether adapting the VAE decoder---via LoRA or full fine-tuning---could reduce the reconstruction gap introduced by the domain shift from natural video to medical dose distributions.

\begin{table}[h]
    \centering
    \caption{Effect of VAE adaptation on GDP--HMM (validation). All variants use the same Any2Any DiT with $v$-pred and 10-step sampling.}
    \label{tab:supp_vae}
    \setlength{\tabcolsep}{6pt}
    \small
    \begin{tabular}{lcc}
        \toprule
        VAE Strategy & MAE$\downarrow$ & Score$\uparrow$ \\
        \midrule
        Frozen VAE (default) & 1.91 & 138.17 \\
        VAE Decoder LoRA & 1.90 & 138.24 \\
        VAE Decoder fine-tuning & 1.89 & 138.39 \\
        Full VAE fine-tuning & 2.54 & 121.76 \\
        \bottomrule
    \end{tabular}
\end{table}

\paragraph{Analysis.}
Decoder-only adaptation yields marginal improvements: LoRA on the decoder reduces MAE by 0.01\,Gy, and full decoder fine-tuning reduces it by 0.02\,Gy while slightly improving the clinical Score. These gains are modest because the Wan VAE's latent space already provides a sufficiently expressive representation, and the DiT backbone compensates for residual distributional differences through its fine-tuned generation process.

In contrast, full VAE fine-tuning (encoder + decoder) dramatically degrades performance (MAE~2.54, Score~121.76). This failure occurs because modifying the encoder destroys the pretrained latent space structure that the DiT backbone relies on, effectively negating the benefit of transfer learning. The encoder-side perturbation causes a distribution mismatch between the latent codes produced during training and those expected by the frozen DiT weights from pretraining.

Based on these results, we adopt the frozen-VAE strategy as the default. It preserves the pretrained latent space, avoids the risk of catastrophic drift from encoder adaptation, and adds no additional training cost. In settings where marginal gains are desired, decoder-only LoRA offers a safe middle ground with negligible overhead.

\subsection{Discussion on MAISI Diffusion Prior}
\label{sec:maisi_ablation}


\begin{table}[t]
    \centering
    \small
    \caption{GDP--HMM head-and-neck results on the MAISI backbone with different output parameterizations. MAE is reported in Gy. Infer time is only reported on deep learning backbone forward without data loading.}
    \label{tab:maisi_param} 
    \begin{tabular}{lcccc}
        \toprule
        Method & Valid MAE & Test MAE & Infer time (H100) & Train time (H100) \\
        \midrule
        Challenge Winner & 2.03 & 2.07 & $<$1\,s & 144\,h \\
        Ours (MAISI, $x_0$-pred, 1 step)          & \textbf{1.89} & \textbf{1.95} & $<$1\,s & 12\,h \\
        Ours (MAISI, noise-pred, 5 steps) & 10.475 & 11.945 & 1.2\,s & 12\,h \\
        Ours (MAISI, noise-pred, 50 steps)& 10.636 & 11.969 & 5.7\,s & 12\,h \\
        \bottomrule
    \end{tabular}
\end{table}

MAISI \cite{guo2025maisi} is originally trained to generate CT images, which is fundamentally different from the task of generating dose. To evaluate whether MAISI can serve as a viable backbone for dose prediction, we ported our training pipeline to the public MAISI latent diffusion model, replacing the Wan VAE+DiT with the MAISI VAE+UNet backbone while keeping the same data preprocessing and optimization schedule as DiffKT3D. Table~\ref{tab:maisi_param} reports performance on the official validation and test sets.

With an $x_0$-prediction objective and joint fine-tuning of MAISI VAE decoder, our re-implementation improves MAE over challenge top1 model while using substantially less training time and without task-specific model crafting. \textit{This provides independent evidence beyond main Wan-based experiments that diffusion priors learned from a large-gap source domain can transfer effectively to a target domain.}

However, we find that switching the same MAISI backbone to a noise-prediction objective while freezing the VAE decoder causes performance to collapse: both validation and test MAE degrade by over \(5\times\), Increasing the number of sampling steps from 5 to 50 does not recover performance (Table~\ref{tab:maisi_param}).

Our closer look indicates that the failure is driven by precision rather than optimization. After standardizing doses, the ground-truth dose maps are normalized to a fixed \([0,1]\) range and then linearly rescaled to \([0,70]\)~Gy. In contrast, the decoded dose from noise-predicted MAISI latents often occupies a wider and case-dependent range, roughly \([0,1.05]\)--\([0,1.2]\), corresponding to \([0,74]\)--\([0,84]\)~Gy after rescaling. In other words, the maximum of the decoded range is no longer pinned at 1 but drifts between about 1.05 and 1.2 across patients. This drifting dynamic range makes it difficult for a single regression head to align voxel intensities across patients, especially near the clinically important 60--70~Gy region, and explains why MAISI behaves well for $x_0$-prediction with a jointly trained decoder but degrades sharply for $v$- or noise-parameterizations under a frozen-decoder regime.

Since DiffKT3D is explicitly designed around a frozen VAE and Any2Any conditioning, we therefore adopt the Wan~2.1 VAE and DiT backbone \cite{wan2025wan,peebles2023dit} rather than MAISI, and we included MAISI-based model as a baseline for DiffKT3D.

\newpage
\section{Statistical Significance Analysis}
\label{sec:statistical_analysis}

To quantify whether the performance improvements of DiffKT3D over the challenge top-1 baseline are statistically significant, we conduct paired $t$-tests on per-patient MAE and Scorecard values across the GDP--HMM test set.

\paragraph{Setup.}
For each of the 498 test patients, we compute: (i)~the voxelwise MAE within the body mask, and (ii)~the clinical Scorecard value aggregating PTV coverage and OAR sparing metrics. We then perform two-sided paired $t$-tests comparing our final model (Any2Any + ScardNFT) against the challenge top-1 regression baseline.

\paragraph{Results.}
Both tests yield $p < 10^{-3}$, confirming that the improvements in MAE (2.07~$\to$~1.93\,Gy) and Scorecard (134.81~$\to$~137.55) are statistically significant and not attributable to random variation across patients.

\paragraph{Discussion on single-step vs.\ multi-step inference.}
While the aggregated Scorecard difference between single-step $v$-pred (133.59) and 10-step $v$-pred (138.17) may appear moderate in absolute terms, the Scorecard aggregates metrics across 30+ regions of interest (ROIs). Many organs distant from the tumor contribute similar scores regardless of sampling depth, which can mask substantial local improvements. For clinically critical structures near the target---where precise dose gradients directly impact treatment quality---the per-organ score differences can be substantial (e.g., differences of 0--12 points for individual OARs). This observation supports the use of multi-step refinement in clinical deployment, where localized dosimetric accuracy in high-gradient regions is paramount.
 \fi

\end{document}